\documentclass[a4paper,twoside]{article}

\usepackage{epsfig}
\usepackage{subcaption}
\usepackage{calc}
\usepackage{amssymb}
\usepackage{amstext}
\usepackage{amsmath}
\usepackage{amsthm}
\usepackage{multicol}
\usepackage{pslatex}
\usepackage{apalike}
\usepackage{algorithm2e}
\usepackage[bottom]{footmisc}
\usepackage{SCITEPRESS}     
\usepackage{graphicx}
\usepackage{hyperref}

\begin{document}

\title{Segmentation-Guided Neural Radiance Fields\\for Novel Street View Synthesis}
\author{\authorname{Yizhou Li\sup{1}, Yusuke Monno\sup{1}, Masatoshi Okutomi\sup{1}, Yuuichi Tanaka\sup{2}, Seiichi Kataoka\sup{3},\\and Teruaki Kosiba\sup{4}}
\affiliation{\sup{1}Institute of Science Tokyo, Tokyo, Japan}
\affiliation{\sup{2}Micware Mobility Co., Ltd., Hyogo, Japan}
\affiliation{\sup{3}Micware Automotive Co., Ltd., Hyogo, Japan}
\affiliation{\sup{4}Micware Navigations Co., Ltd, Hyogo, Japan}
\email{\{yli,ymonno\}@ok.sc.e.titech.ac.jp, mxo@ctrl.titech.ac.jp, \{tanaka\_yuu,kataoka\_se,kosiba\_te\}@micware.co.jp}
}


\abstract{Recent advances in Neural Radiance Fields (NeRF) have shown great potential in 3D reconstruction and novel view synthesis, particularly for indoor and small-scale scenes. However, extending NeRF to large-scale outdoor environments presents challenges such as transient objects, sparse cameras and textures, and varying lighting conditions. In this paper, we propose a segmentation-guided enhancement to NeRF for outdoor street scenes, focusing on complex urban environments. Our approach extends ZipNeRF and utilizes Grounded SAM for segmentation mask generation, enabling effective handling of transient objects, modeling of the sky, and regularization of the ground. We also introduce appearance embeddings to adapt to inconsistent lighting across view sequences. Experimental results demonstrate that our method outperforms the baseline ZipNeRF, improving novel view synthesis quality with fewer artifacts and sharper details.}

\onecolumn \maketitle \normalsize \setcounter{footnote}{0} \vfill

\begin{figure*}[!t]
    \centering
\includegraphics[width=1.0\linewidth]{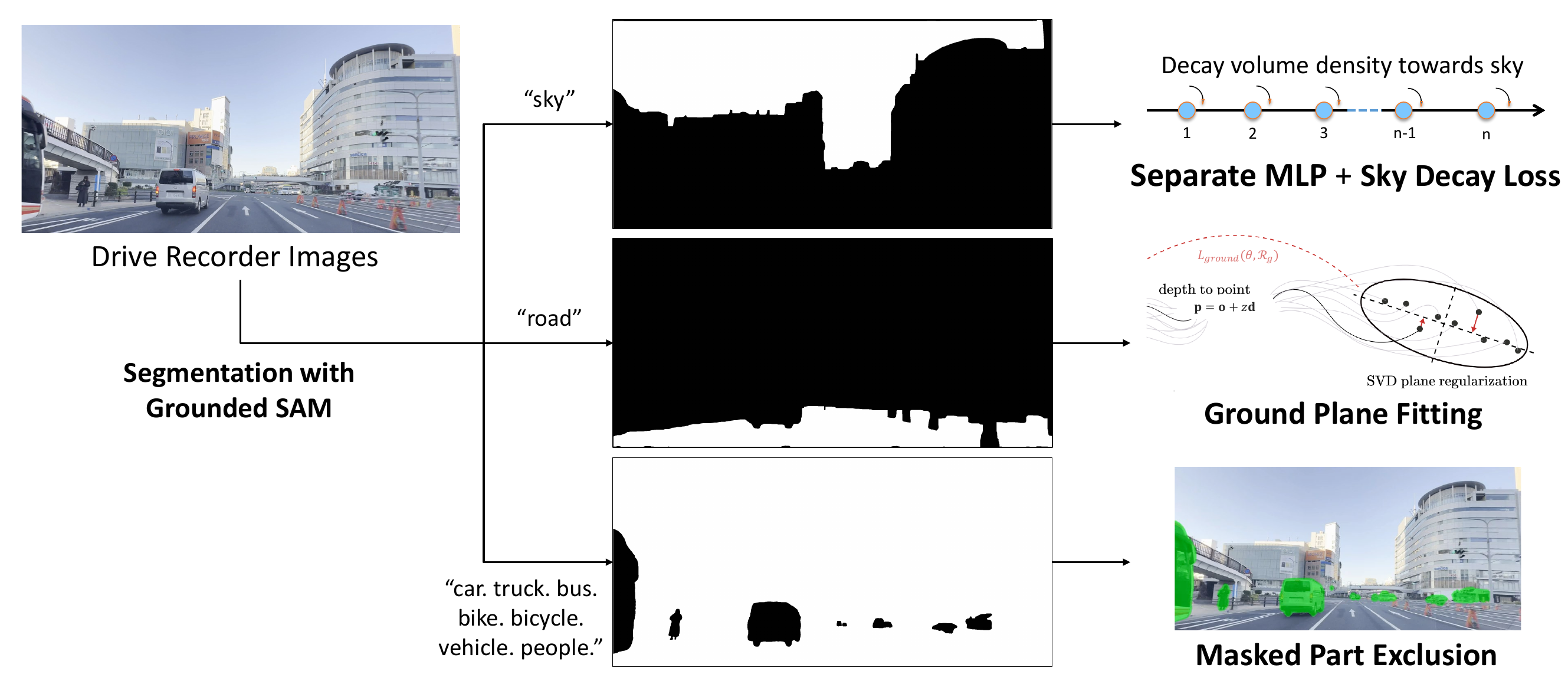}
    \caption{The overview of our approaches for different segmentation regions.}
\label{fig:mask}
\end{figure*}

\section{\uppercase{Introduction}}

Neural Radiance Fields (NeRF)~\cite{nerf} have emerged as a powerful tool for reconstructing 3D scenes and generating novel view images with impressive quality, offering significant potential for applications such as autonomous driving and augmented reality. Although NeRF performs well in bounded scenes, extending it to unbounded outdoor scenes such as urban street scenes presents unique challenges. While various methods have been proposed to tackle different challenges in outdoor scenes ~\cite{nerf++,mipnerf360,blocknerf,urf,suds}, a unified framework to address these challenges is still in the developing phase.

In this paper, we present an enhanced method of NeRF specifically tailored for novel view synthesis (NVS) of street views. Our method is based on ZipNeRF~\cite{zipnerf}, one of the grid-based variants of NeRFs~\cite{zipnerf,ingp,dvgo} known for its improved efficiency and quality. We extend it to address the challenges associated with outdoor scenarios.

Specifically, we focus on NVS of outdoor street scenes using monocular video clips captured by a video recorder mounted on a car. This is inherently challenging due to the dynamic nature of transient objects such as vehicles and pedestrians, the presence of sparse textures in certain regions such as the sky and the ground, and the variations in lighting conditions across different video clips. We summarize these challenges as follows.

First, transient objects such as vehicles and pedestrians present significant challenges, as they disrupt the consistency across video frames required for accurate NeRF learning. Second, the sky often results in erroneous near-depth estimation due to the lack of textures and defined features, which leads to floating artifacts during NVS. Third, limited textures in the ground often lead to poor geometry estimation, producing noticeable artifacts during NVS. Finally, street view videos captured at different times introduce inconsistent lighting conditions, which contradict NeRF's assumption of consistent colors across views. These inconsistencies result in blurry and inaccurate NVS.

To overcome these challenges, we leverage the semantic information from Grounded SAM~\cite{grounded_sam}, a semantic segmentation model based on SAM~\cite{grounding_dino} and Grounding DINO~\cite{grounding_dino} that utilizes text prompts, to obtain precise segmentation masks for each target. Concretely, we obtain the masks of transient objects, sky, and ground using Grounded SAM, as shown in Fig.~\ref{fig:mask}. We then apply different techniques to handle each segmented region as follows.

\begin{itemize}

\item \textbf{Transient objects:} We mask them out during the training, effectively excluding them from contributing to the learned densities and colors, which reduces NVS artifacts.

\item \textbf{Sky:} We utilize a separate sky-specific hash representation~\cite{suds} that estimates the sky's appearance based solely on view direction, ensuring accurate background representation without causing erroneous density in the foreground. Additionally, we implement a sky decay loss to further suppress artifacts resulting from the sky region.

\item \textbf{Ground:} We introduce a plane-fitting regularization loss in PlaNeRF~\cite{planerf} that encourages the ground surface to conform to a planar geometry, thus enhancing the reconstruction quality of ground areas.

\item \textbf{Inconsistent lighting:} We adopt an appearance embedding strategy inspired by BlockNeRF~\cite{blocknerf} and URF~\cite{urf}. This allows us to learn an image-wise appearance embedding that captures different lighting conditions, enabling our model to disentangle irradiance from appearance and produce consistent colors across the scene.
\end{itemize}

We evaluate our improved method to our real data, consisting of 12 video clips entering and exiting an intersection from different directions. The results demonstrate substantial improvements over the baseline ZipNeRF, particularly in reducing artifacts.

\begin{figure*}[!h]
    \centering
\includegraphics[width=1.0\linewidth]{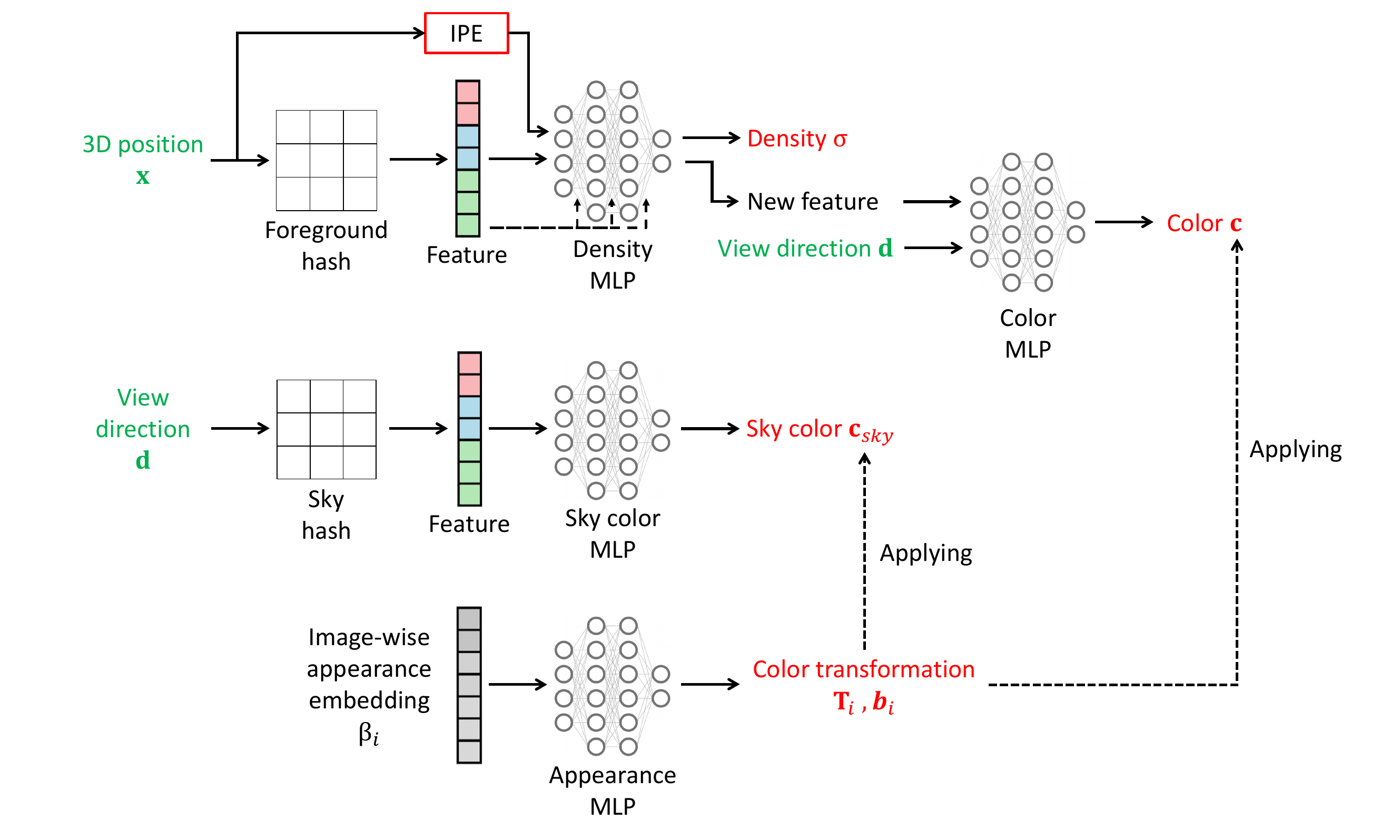}
    \caption{The overview of our network architecture.}
\label{fig:network}
\end{figure*}

\section{\uppercase{Methodology}}

In this section, we describe our methodology, including the definitions of symbols and the processes applied to handle different segmented regions in our segmentation-guided NeRF enhancement.

\subsection{NeRF Formulation}

Let $\{I_1, I_2, \dots, I_N\}$ represent a set of input images captured from different viewpoints using a monocular camera, where $ N $ is the total number of images. Our goal is to reconstruct a 3D scene representation and generate novel view images of the scene. The neural radiance field is represented as a combination of grid-based features and a Multi-Layer Perceptron~(MLP). The output of our method can be represented by the following function: \begin{equation} f(\mathbf{x}, \mathbf{d}) \rightarrow (\mathbf{c}, \sigma), \end{equation} where $ f(\mathbf{x}, \mathbf{d}) $ takes a 3D point $ \mathbf{x} \in \mathbb{R}^3 $ and a view direction $ \mathbf{d} \in \mathbb{R}^3 $ as the inputs, and returns the corresponding color $ \mathbf{c} \in \mathbb{R}^3 $ and density $ \sigma \in \mathbb{R} $.

\subsection{Segmentation Masks}

We utilize Grounded SAM~\cite{grounded_sam} to segment the input images and obtain three regions: transient objects, sky, and ground. For the segmentation of each region, we use specific text prompts, as shown in Fig.~\ref{fig:mask}, and obtain the following masks:

\begin{itemize} \item $ M_{t} $: Mask for transient objects, \item $ M_{s} $: Mask for the sky region, \item $ M_{g} $: Mask for the ground region, \end{itemize}
where each mask $ M \in \{0, 1\} $ is a binary mask that indicates whether a pixel belongs to the specified region ($ M = 1 $) or not ($ M = 0 $).

\subsection{Handling Different Regions}

\subsubsection{Transient Objects}

For the transient objects, represented by the mask $ M_t $, we exclude their contributions during the NeRF training phase. Specifically, we set the loss contribution from the rays reaching the transient objects to zero. The color reconstruction loss $L_{\text{rgb}}$ is defined as \begin{equation} L_{\text{rgb}}(\theta) = \sum_{i} E_{r \sim I_i} \left[ (1 - M_t(r)) \cdot \left| \left| C(r; \beta_i) - C_{\text{gt}, i}(r) \right| \right|_2^2  \right], \end{equation} where $\theta$ is the network parameters, $ C(r; \beta_i) $ is the predicted color of ray $ r $ with appearance compensation parameter $ \beta_i $ (detailed in Sec.~\ref{subsec:appearance_embedding}), and $C_{\text{gt}, i}(r)$ is the ground truth color of ray $ r $ in image $ i $.

\subsubsection{Sky Region}

For the sky region, represented by the mask $ M_s $, we adopt a separate sky-specific representation for modeling the sky's appearance as shown in Fig.~\ref{fig:network}. The sky's color is estimated based solely on the view direction $ \mathbf{d} $, as the sky can be considered infinitely far away. We blend the sky representation with the foreground using an alpha map derived from the accumulated density along each ray: \begin{equation} C(r; \beta_i) = \int_{t_n}^{t_f} w(t) \cdot \Gamma(\beta_i) \cdot \mathbf{c}(t) dt + \mathbf{c}_{sky}(\mathbf{d}), \end{equation} where $ \Gamma(\beta_i) $ is the appearance compensation transformation, $ \mathbf{c}_{sky}(\mathbf{d}) $ is the sky color predicted by the sky network, and $ w(t) $ represents the volume rendering weight defined as \begin{equation} w(t) = \exp \left( - \int_{t_n}^t \sigma(s) ds \right) \cdot \sigma(t). \end{equation} 

We then introduce the sky decay loss $ L_{\text{sky}} $ to both suppress density estimates in the sky region and enhance the accumulated density of rays not marked by the sky mask. This approach helps to prevent the generation of artifacts such as floaters and ensures a more accurate representation of non-sky regions. The sky decay loss is defined as
\begin{align}
    L_{\text{sky}}(\theta) &= \sum_{i} E_{r \sim I_i} \left[ M_s(r) \int_{t_n}^{t_f} w(t)^2 \, dt \right] \nonumber \\
    &\quad - \sum_{i} E_{r \sim I_i} \left[ (1 - M_s(r)) \int_{t_n}^{t_f} w(t)^2 \, dt \right],
\end{align}
where the first term suppresses the density estimates for rays marked by the sky mask ($M_s(r) = 1$), and the second term enhances the accumulated density for rays not marked by the sky mask ($M_s(r) = 0$).

While applying the sky decay loss, we observed unintended side effects, particularly due to hash collisions inherent in the grid-based ZipNeRF. These collisions could cause the decay of densities in non-sky foreground regions where suppression is not intended. To mitigate these negative effects, we incorporate positional embeddings of 3D point coordinates $\mathbf{x}$ as additional inputs to the MLP, as shown in Fig.~\ref{fig:network}. This ensures that the density estimation relies not only on the hash features but also on accurate spatial information from $\mathbf{x}$. To prevent the MLP from over-relying on the 3D coordinates, we also add residual connections~\cite{resnet}, allowing the hash features to be directly fed into the intermediate layers of the MLP.
The final MLP input is given by
\begin{equation} \mathbf{z} = [f_{\text{hash}}(\mathbf{x}), \gamma(\mathbf{x})], \end{equation}
where $f_{\text{hash}}(\mathbf{x})$ is the feature obtained from the hash, and $\gamma(\mathbf{x})$ is the integrated positional embedding (IPE)~\cite{mipnerf} of $\mathbf{x}$.

\subsubsection{Ground Region}

For the ground region, represented by the mask $ M_g $, we adopt a plane regularization method based on Singular Value Decomposition (SVD) following PlaNeRF~\cite{planerf} as shown in Fig.~\ref{fig:mask}. We apply this regularization to ensure that the predicted points on the ground conform to a planar structure, which helps in achieving a more consistent reconstruction of the ground surface.

Given a patch of rays $ \mathcal{R}_g $ for the ground region, we define the predicted point cloud 
\begin{equation}
    \mathcal{P} = \{ \mathbf{p}_r = \mathbf{o}_r + z_r \mathbf{d}_r \mid r \in \mathcal{R}_g \}
\end{equation}
where $ \mathbf{o}_r $ is the ray origin, $ z_r $ is the rendered depth, and $ \mathbf{d}_r $ is the ray direction of ray $ r $. The least-squares plane defined by a point $ \hat{\mathbf{p}}_c $ and a normal unitary vector $ \mathbf{n} $ is obtained by solving the following optimization problem: 
\begin{equation}
    \min_{\substack{\hat{\mathbf{p}}_c, \mathbf{n}}} \sum_{r \in R} \left( (\mathbf{p}_r - \hat{\mathbf{p}}_c) \cdot \mathbf{n} \right)^2,
\end{equation}
where the point $ \hat{\mathbf{p}}_c $ is the barycenter of the point cloud: 
\begin{equation}
    \hat{\mathbf{p}}_c = \frac{1}{N_p} \sum_{r \in \mathcal{R}_g} \mathbf{p}_r,
\end{equation}
where $ N_p $ is the number of points in the patch.
We form a matrix $ \mathbf{A} $ from the differences between each point and the barycenter as
\begin{equation}
   \mathbf{A} = \begin{bmatrix} \mathbf{p}_0 - \hat{\mathbf{p}}_c & \mathbf{p}_1 - \hat{\mathbf{p}}_c & \cdots & \mathbf{p}_{N_p} - \hat{\mathbf{p}}_c \end{bmatrix}^T.
\end{equation}

The plane normal $ \mathbf{n} $ is given by the right singular vector corresponding to the smallest singular value of $ \mathbf{A} $, which can be found using SVD. We regularize the NeRF-rendered points to this plane by minimizing the smallest singular value $ \sigma_3 $ of $ \mathbf{A} $ as \begin{equation} L_{\text{ground}}(\theta, \mathcal{R}_g) = \sigma_3(\theta, \mathcal{R}_g). \end{equation}

This regularization encourages the points in the ground region to lie on a plane, thereby improving the geometry of the reconstructed ground.

\subsection{Appearance Embedding for Lighting Inconsistencies}
\label{subsec:appearance_embedding}

To address lighting inconsistencies across video clips, we follow URF~\cite{urf} to perform an affine mapping of the radiance predicted by the shared network as shown in Fig.~\ref{fig:network}. This affine transformation is represented by a $3 \times 3$ matrix and a $1 \times 3$ shift vector, both of which are decoded from a per-image latent code $ \beta_i \in \mathbb{R}^B $ as
\begin{equation} \Gamma(\beta_i) = (\mathbf{T}_i, \mathbf{b}_i): \mathbb{R}^B \to (\mathbb{R}^{3 \times 3}, \mathbb{R}^{1 \times 3}), \end{equation}
where $\mathbf{T}_i$ represents the color transformation matrix and $\mathbf{b}_i$ represents the shift vector.

The color transformation for the radiance $\mathbf{c}$ predicted by the network is then performed as
\begin{equation} \mathbf{c}' = \mathbf{T}_i \mathbf{c} + \mathbf{b}_i, \end{equation}
where $\mathbf{c} \in \mathbb{R}^3$ is the original radiance, and $\mathbf{c}' \in \mathbb{R}^3$ is the transformed color.

This affine mapping models lighting and exposure variations with a more restrictive function, thereby reducing the risk of unwanted entanglement when jointly optimizing the scene radiance parameters $ \theta $ and the appearance mappings $ \beta $.

\subsection{Overall Loss Function}

The complete loss function $L_{\text{total}}$ is a weighted combination of the above-mentioned components, which is described as

\begin{equation}
    L_{\text{total}} = L_{\text{rgb}} + \lambda_{\text{sky}} L_{\text{sky}} + \lambda_{\text{ground}} L_{\text{ground}}.
\end{equation}

\begin{figure*}[!t]
    \centering
\includegraphics[width=0.69\linewidth]{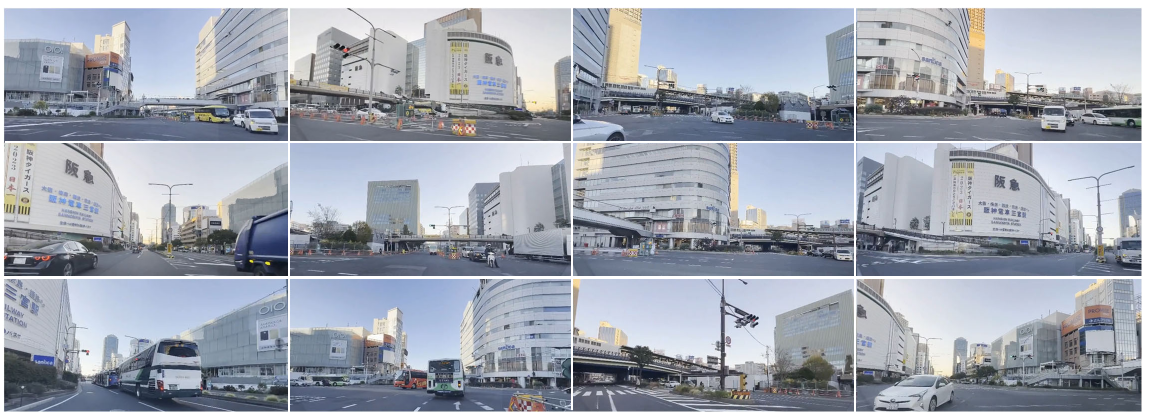}
\includegraphics[width=0.29\linewidth]{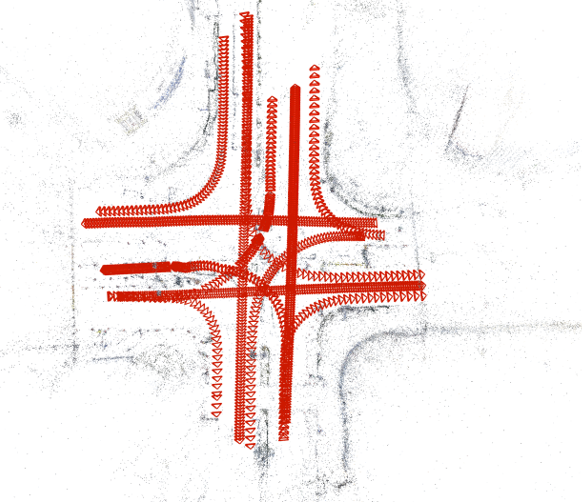}
    \caption{(Left) Sample images of our dataset for an intersection. Each image is from each of 12 video clips. (Right) Estimated camera poses by using COLMAP.}
\label{fig:dataset}
\end{figure*}

\begin{figure*}[!t]
    \centering
\includegraphics[width=1.0\linewidth]{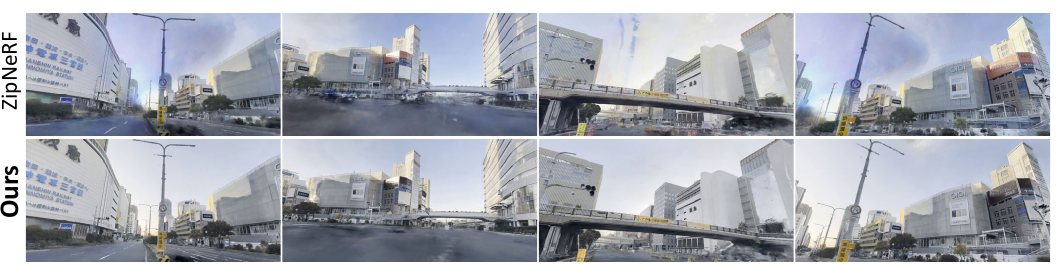}
    \caption{Comparison of ZipNeRF and the proposed method for novel view synthesis.}
\label{fig:comparison}
\end{figure*}

\section{\uppercase{Experiments}}

In this section, we present the experimental setups, including implementation and dataset details. Then, we present a quantitative comparison of our proposed method with the baseline ZipNeRF to demonstrate the effectiveness of our segmentation-guided NeRF enhancement for outdoor street scenes.

\subsection{Dataset}

We collected our dataset using an iPhone 15, capturing video footage of the Sannomiya intersection in Kobe, Japan. The dataset includes a total of 12 video clips, as shown in Fig.~\ref{fig:dataset}: four straight directions (South to North, North to South, West to East, and East to West) and eight turning directions (e.g., South to West, South to East, etc.). As shown in Fig.~\ref{fig:dataset}, each video clip was captured throughout different times of the day, thus with varying lighting conditions.

After capturing the videos, we extracted a total of 1,112 frames from the 12 video clips. The original frames are in 4K resolution ($3840 \times 2160$). Since the videos were recorded from inside a car, we cropped out the windshield area, resulting in a reduced resolution of $3376 \times 1600$. To make the data more manageable for processing, we further resized the frames to $2110 \times 1100$, which were then used for COLMAP and NeRF training.

After resizing, we used COLMAP~\cite{schoenberger2016sfm}, an open-source Structure-from-Motion (SfM) tool, to obtain the camera poses and intrinsic parameters for each frame. The estimated intrinsic parameters and camera poses were then used for our NeRF training. 

This dataset presents significant challenges, including changing lighting conditions, transient objects (e.g., pedestrians and vehicles), and sparse textures in the sky and ground regions, making it ideal for evaluating the robustness of our proposed method.

\subsection{Implementation Details}

Our implementation is based on PyTorch and an NVIDIA RTX 4090 GPU. The total training time for our data was approximately $6$ hours. We used the Adam optimizer for parameter updates, with a batch size of $4096$ rays per iteration. The initial learning rate was set to $0.01$ and gradually decayed to $0.001$ using a cosine annealing schedule over a maximum of $50,000$ iterations. The loss weights were set as $\lambda_{\text{sky}} = \lambda_{\text{ground}} = 0.0001$.

\begin{figure*}[!t]
    \centering
\includegraphics[width=1.0\linewidth]{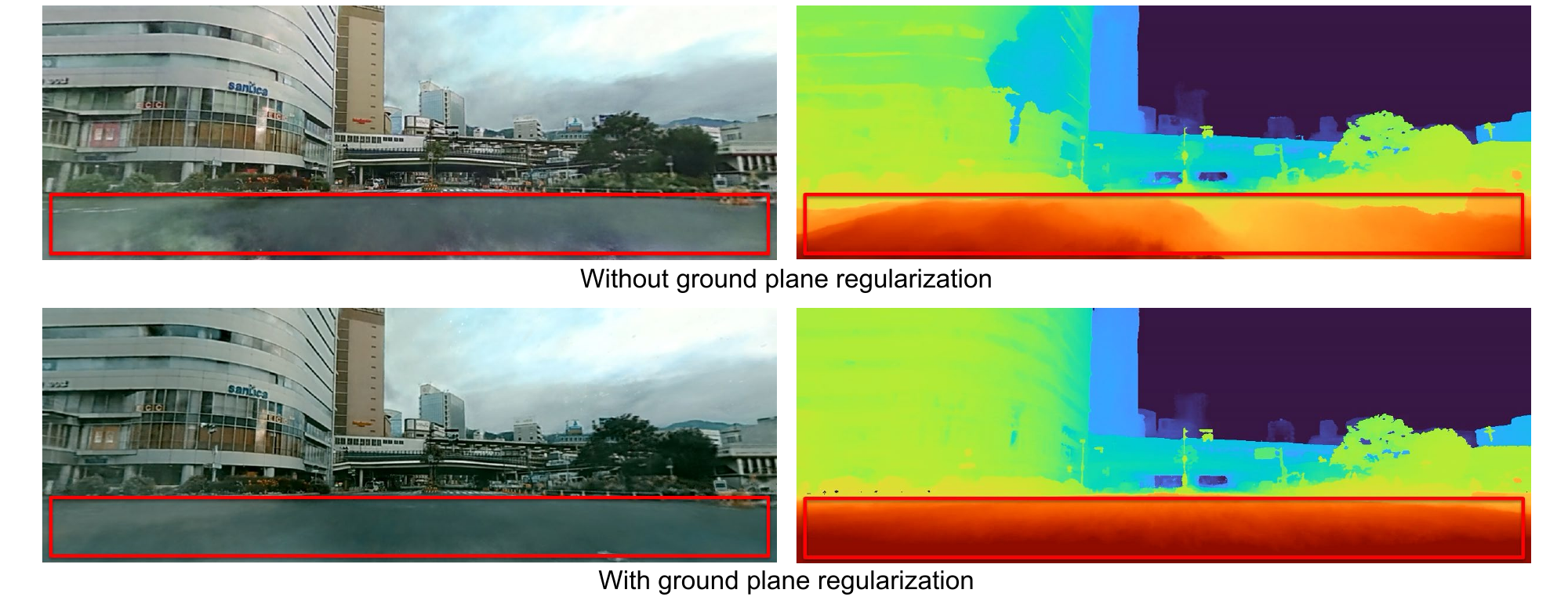}
    \vspace{-5mm}
    \caption{Comparison of the cases without and with ground plane regularization.}
\label{fig:ground}
\end{figure*}

\begin{figure*}[!t]
    \centering
\includegraphics[width=1.0\linewidth]{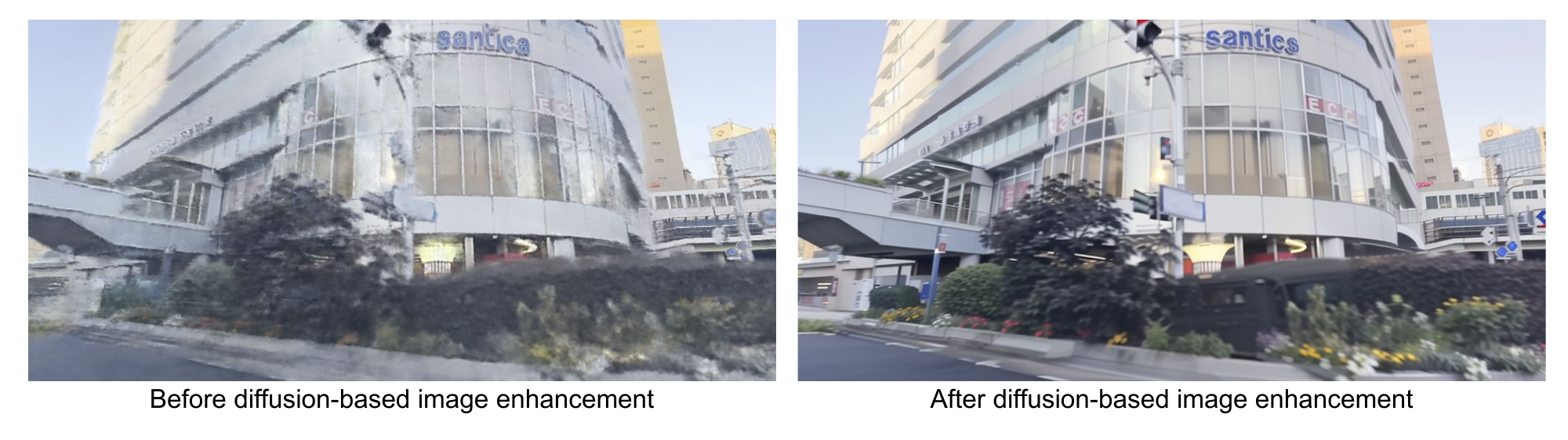}
    \vspace{-6mm}
    \caption{Comparison of the results before and after our diffusion-based image enhancement.}
\label{fig:diffusion}
\end{figure*}

\subsection{Qualitative Comparison}

We conducted a quantitative comparison between our proposed method and the baseline ZipNeRF. Figure~\ref{fig:comparison} shows the visualization of the results, where we compare the novel view images of the reconstructed scenes in terms of visual quality and the presence of artifacts. The proposed method demonstrates notable improvements over ZipNeRF, particularly in handling challenging outdoor conditions.

In the images produced by ZipNeRF, we observed significant blurring, color artifacts, and floating artifacts in areas with sparse textures, such as the sky and the ground. In contrast, our method effectively mitigates these issues. By employing segmentation-guided enhancements such as sky-specific modeling, transient object masking, and ground plane regularization, our approach produces clearer details, fewer artifacts, and sharper details in the novel view images.
Figure~\ref{fig:ground} shows the detailed comparison of rendered images and depth maps in the cases without and with ground plane regularization. We can clearly observe that the plane regularization helps to generate more reliable depth maps, reducing floating artifacts in the ground regions.
The above results highlight the effectiveness of our segmentation-guided NeRF enhancements in addressing the unique challenges posed by outdoor street environments.

Even though our method generates appealing results, it still generates artifacts if the rendered novel view is far from the training views, as exemplified in the left image of Fig.~\ref{fig:diffusion}. To enhance the results for those views, we apply our previously proposed diffusion-based image restoration method~\cite{li2025tdm}, where we restore the rendered images with artifacts based on the pre-trained stable diffusion model~\cite{LatentDiffusion} and a fine-tuned ControlNet~\cite{controlnet}. The right image of Fig.~\ref{fig:diffusion} shows the result after our diffusion-based enhancement, demonstrating a visually pleasing result by utilizing the power of a diffusion model.

\section{\uppercase{Conclusion}}

In this paper, we have presented a segmentation-guided NeRF enhancement for novel street view synthesis. Building on ZipNeRF, we have introduced techniques to address challenges like transient objects, sparse textures, and lighting inconsistencies. By utilizing Grounded SAM for segmentation and introducing appearance embeddings, our method effectively handles these challenges. Qualitative results have demonstrated that our method outperforms the baseline ZipNeRF, producing fewer artifacts, sharper details, and improved geometry, especially in challenging areas like the sky and the ground.

\bibliographystyle{apalike}
{\small
\bibliography{VISAPP2025}}

\begin{thebibliography}{}

\bibitem[Barron et~al., 2021]{mipnerf}
Barron, J.~T., Mildenhall, B., Tancik, M., Hedman, P., Martin-Brualla, R., and Srinivasan, P.~P. (2021).
\newblock {Mip-NeRF}: A multiscale representation for anti-aliasing neural radiance fields.
\newblock In {\em Proceedings of the IEEE/CVF International Conference on Computer Vision (ICCV)}, pages 5855--5864.

\bibitem[Barron et~al., 2022]{mipnerf360}
Barron, J.~T., Mildenhall, B., Verbin, D., Srinivasan, P.~P., and Hedman, P. (2022).
\newblock {Mip-NeRF} 360: Unbounded anti-aliased neural radiance fields.
\newblock In {\em Proceedings of the IEEE/CVF Conference on Computer Vision and Pattern Recognition (CVPR)}, pages 5470--5479.

\bibitem[Barron et~al., 2023]{zipnerf}
Barron, J.~T., Mildenhall, B., Verbin, D., Srinivasan, P.~P., and Hedman, P. (2023).
\newblock {Zip-NeRF}: Anti-aliased grid-based neural radiance fields.
\newblock In {\em Proceedings of the IEEE/CVF International Conference on Computer Vision (ICCV)}, pages 19697--19705.

\bibitem[He et~al., 2016]{resnet}
He, K., Zhang, X., Ren, S., and Sun, J. (2016).
\newblock Deep residual learning for image recognition.
\newblock In {\em Proceedings of the IEEE Conference on Computer Vision and Pattern Recognition (CVPR)}, pages 770--778.

\bibitem[Li et~al., 2025]{li2025tdm}
Li, Y., Liu, Z., Monno, Y., and Okutomi, M. (2025).
\newblock {TDM}: Temporally-consistent diffusion model for all-in-one real-world video restoration.
\newblock In {\em Proceedings of International Conference on Multimedia Modeling (MMM)}, pages 155--169.

\bibitem[Liu et~al., 2023]{grounding_dino}
Liu, S., Zeng, Z., Ren, T., Li, F., Zhang, H., Yang, J., Li, C., Yang, J., Su, H., Zhu, J., et~al. (2023).
\newblock {Grounding DINO}: Marrying {DINO} with grounded pre-training for open-set object detection.
\newblock {\em arXiv preprint 2303.05499}.

\bibitem[Mildenhall et~al., 2020]{nerf}
Mildenhall, B., Srinivasan, P.~P., Tancik, M., Barron, J.~T., Ramamoorthi, R., and Ng, R. (2020).
\newblock {NeRF}: Representing scenes as neural radiance fields for view synthesis.
\newblock In {\em Proceedings of European Conference on Computer Vision (ECCV)}, pages 405--421.

\bibitem[M{\"u}ller et~al., 2022]{ingp}
M{\"u}ller, T., Evans, A., Schied, C., and Keller, A. (2022).
\newblock Instant neural graphics primitives with a multiresolution hash encoding.
\newblock {\em ACM Transactions on Graphics (TOG)}, 41(4):1--15.

\bibitem[Rematas et~al., 2022]{urf}
Rematas, K., Liu, A., Srinivasan, P.~P., Barron, J.~T., Tagliasacchi, A., Funkhouser, T., and Ferrari, V. (2022).
\newblock Urban radiance fields.
\newblock In {\em Proceedings of the IEEE/CVF Conference on Computer Vision and Pattern Recognition (CVPR)}, pages 12932--12942.

\bibitem[Ren et~al., 2024]{grounded_sam}
Ren, T., Liu, S., Zeng, A., Lin, J., Li, K., Cao, H., Chen, J., Huang, X., Chen, Y., Yan, F., Zeng, Z., Zhang, H., Li, F., Yang, J., Li, H., Jiang, Q., and Zhang, L. (2024).
\newblock {Grounded SAM}: Assembling open-world models for diverse visual tasks.
\newblock In {\em arXiv preprint 2401.14159}.

\bibitem[Rombach et~al., 2022]{LatentDiffusion}
Rombach, R., Blattmann, A., Lorenz, D., Esser, P., and Ommer, B. (2022).
\newblock High-resolution image synthesis with latent diffusion models.
\newblock In {\em Proceedings of the IEEE/CVF Conference on Computer Vision and Pattern Recognition (CVPR)}, pages 10684--10695.

\bibitem[Sch\"{o}nberger and Frahm, 2016]{schoenberger2016sfm}
Sch\"{o}nberger, J.~L. and Frahm, J.-M. (2016).
\newblock Structure-from-motion revisited.
\newblock In {\em Proceedings of IEEE Conference on Computer Vision and Pattern Recognition (CVPR)}, pages 4104--4113.

\bibitem[Sun et~al., 2022]{dvgo}
Sun, C., Sun, M., and Chen, H.-T. (2022).
\newblock Direct voxel grid optimization: Super-fast convergence for radiance fields reconstruction.
\newblock In {\em Proceedings of the IEEE/CVF Conference on Computer Vision and Pattern Recognition (CVPR)}, pages 5459--5469.

\bibitem[Tancik et~al., 2022]{blocknerf}
Tancik, M., Casser, V., Yan, X., Pradhan, S., Mildenhall, B., Srinivasan, P.~P., Barron, J.~T., and Kretzschmar, H. (2022).
\newblock {Block-NeRF}: Scalable large scene neural view synthesis.
\newblock In {\em Proceedings of the IEEE/CVF Conference on Computer Vision and Pattern Recognition (CVPR)}, pages 8248--8258.

\bibitem[Turki et~al., 2023]{suds}
Turki, H., Zhang, J.~Y., Ferroni, F., and Ramanan, D. (2023).
\newblock {SUDS}: Scalable urban dynamic scenes.
\newblock In {\em Proceedings of the IEEE/CVF Conference on Computer Vision and Pattern Recognition (CVPR)}, pages 12375--12385.

\bibitem[Wang et~al., 2024]{planerf}
Wang, F., Louys, A., Piasco, N., Bennehar, M., Rold{\~a}ao, L., and Tsishkou, D. (2024).
\newblock {PlaNeRF}: {SVD} unsupervised {3D} plane regularization for {NeRF} large-scale urban scene reconstruction.
\newblock In {\em Proceedings of International Conference on 3D Vision (3DV)}, pages 1291--1300.

\bibitem[Zhang et~al., 2020]{nerf++}
Zhang, K., Riegler, G., Snavely, N., and Koltun, V. (2020).
\newblock {NeRF++}: Analyzing and improving neural radiance fields.
\newblock {\em arXiv preprint 2010.07492}.

\bibitem[Zhang et~al., 2023]{controlnet}
Zhang, L., Rao, A., and Agrawala, M. (2023).
\newblock Adding conditional control to text-to-image diffusion models.
\newblock In {\em Proceedings of the IEEE/CVF International Conference on Computer Vision (ICCV)}, pages 3836--3847.

\end{thebibliography}

\end{document}